\documentclass{vgtc} 

\graphicspath{{figs/}{figures/}{./}}

\usepackage{times}

\usepackage{booktabs}
\usepackage{xspace}
\usepackage{enumitem}
\usepackage{mathptmx}
\usepackage{amsmath,amssymb}
\usepackage{graphicx}
\usepackage{subcaption}
\usepackage{xcolor}
\usepackage{multirow}
\usepackage{accessibility}
\usepackage{accessibility}
\usepackage{float}
\usepackage[table]{xcolor}
\onlineid{0}
\vgtccategory{Research}
\vgtcinsertpkg

\title{VEIL: How Visual Encoding Hijacking Induces Bias In Vision Models}

\hypersetup{
  pdftitle={VEIL: How Visual Encoding Hijacking Induces Bias in Vision Models},
  pdfauthor={Anonymous Authors},
  pdfsubject={Visualization and time-series classification},
  pdfkeywords={visualization, time-series classification, visual encoding, machine perception, reproducibility}
}

\author{
\parbox{\textwidth}{\centering
Suranjana Sooraj\thanks{Emails: \{ssooraj, lanche, mvenkat, dyuliu\}@ucdavis.edu} \quad
Xuyang Chen\footnotemark[1] \quad
Madhumitha Venkatesan\footnotemark[1] \quad
Dongyu Liu\footnotemark[1]\\[0.3em]
\scriptsize University of California, Davis
}
}

\usepackage{xcolor}
\usepackage{marginnote}

\abstract{
Rendering time series as chart images for CNN-based classification has become increasingly common in time-series classification (TSC). However, it remains unclear whether models learn underlying temporal patterns or rely on encoding-specific visual cues introduced by chart design. We present VEIL: a systematic study examining how chart encodings influence learned representations through complementary analyses of similarity, transferability, and attribution. Attention-guided training appears to mitigate this effect when encoding sensitivity is consistently identified across diagnostics, but provides limited or negative benefit when such signals are absent. These findings position VEIL within the broader question of how machines perceive visualizations---extending graphical perception from human readers to vision models---and show that visualization design choices shape learned representations in ways that warrant treating chart-based TSC as a representation and measurement problem rather than a simple modeling decision.
}


\begin{document}
\maketitle
\maketitle
\section{Introduction}
\label{sec:introduction}
Building on the widespread adoption of deep learning for TSC~\cite{fawaz2019deep, foumani2024survey}, recent work has increasingly rendered numerical sequences as images and processed them with convolutional backbones~\cite{hatami2018classification, li2023vitst, oh2025tssi, wang2015imaging}. This visual formulation is motivated by the opportunity to reuse powerful vision backbones, including pretrained models, and to make model behavior more inspectable by overlaying attribution methods such as Grad-CAM \cite{selvaraju2017gradcam} on chart images. These approaches have achieved competitive performance, suggesting that chart representations can bridge time-series data and visual learning models~\cite{wu2022timesnet, ni2025harnessing}. However, this raises a key question: what do models actually learn from these visual representations? While visual encodings preserve temporal structure, they also introduce encoding-specific patterns (e.g., stroke continuity, edge boundaries, point density) not inherent to the signal. Models may thus rely on these visual artifacts rather than temporal dynamics~\cite{geirhos2020shortcut, geirhos2019texture}, creating an illusion of learning where they appear to capture meaningful patterns but may instead exploit superficial cues.

%

We term this \emph{visual encoding hijacking}: encoding-dependent behavior where representations align more with rendering style than temporal class structure, distinct from legitimate encoding-specific evidence (e.g., line continuity, bar boundaries, scatter density) that faithfully reflects signal properties. Hijacking is diagnosed jointly through low cross-encoding alignment, weak transfer, and attribution shifts—not any single metric. This effect eludes standard accuracy and varies across encodings and datasets. Despite the growing use of image-based TSC, it remains unclear whether models learn encoding-invariant representations or shortcuts. We investigate: (RQ1) do different visual encodings produce consistent or divergent learned representations of the same time series? (RQ2) do features learned from one encoding transfer across encodings, or are they encoding-specific? and (RQ3) what visual evidence do CNN-based TSC models rely on across encodings and datasets, and when can attention guidance reduce encoding-sensitive behavior? To address these, we introduce VEIL\footnote{\textbf{V}isual \textbf{E}ncoding \textbf{I}llusion in \textbf{L}earning.}, a diagnostic framework using representation similarity (CKA \cite{kornblith2019similarity}), cross-encoding transfer via linear probing, and attribution (Grad-CAM \cite{selvaraju2017gradcam}), along with geometric (PCA \cite{jolliffe2002principal}) and separability (UMAP \cite{alain2017probing}) analyses, plus Encoding Sensitivity Index (ESI). We evaluate four encodings (line, area, bar, scatter) across 31 UCR datasets \cite{dau2019ucr, venkatesan2026vtbenchmultimodalframeworktimeseries}. As graphical perception studies the human ability to read charts, VEIL examines how models read them, framing encoding choice as a measurement decision rather than a tooling preference \cite{lee2024assessing}.


Our findings reveal substantial encoding effects, with limited cross-encoding transfer in most cases, indicating encoding-specific feature learning. Attribution and perturbation analyses show sensitivity to rendering-dependent cues, though this does not establish whether such cues are shortcuts or legitimate encoding-specific evidence reflecting true signal structure. 
VEIL\footnote{\url{https://anonymous.4open.science/r/VEIL/}} underscores encoding choice as a key inductive bias in image-based TSC. Our contributions are as follows:
\begin{itemize}[leftmargin=10pt]
\item We formalize \emph{visual encoding hijacking}---an encoding-dependent behavior that undermines cross-encoding representation alignment and transfer---and distinguish it from legitimate encoding-specific evidence that reflects genuine signal structure.
\vspace{-0.7em}
\item We propose VEIL: a diagnostic framework combining representation similarity, transferability, and attribution analysis to systematically evaluate encoding effects.
\vspace{-0.7em}
\item We present an empirical study across 31 datasets showing that encoding choice is associated with ample variation in representation alignment, cross-encoding transfer, and model attention.
\end{itemize}

\section{Related Work} 
\label{sec:related_work}
\begin{figure*}[!t]
  \centering
  \includegraphics[width=\textwidth]{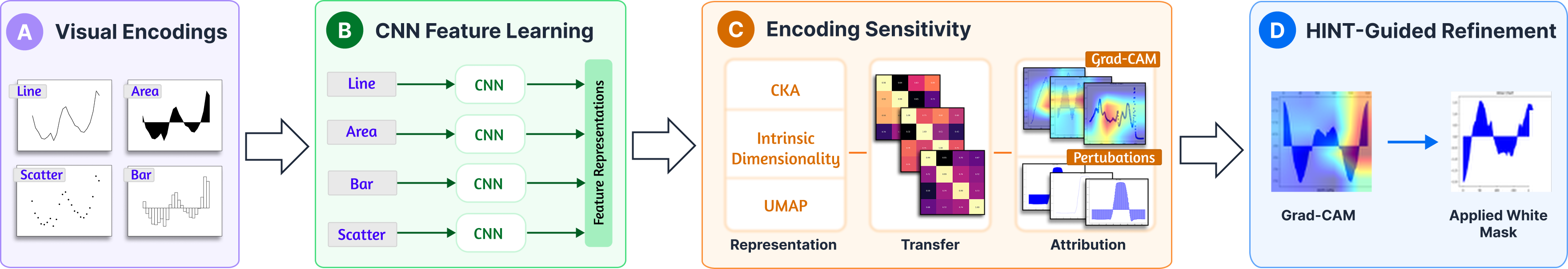}
   \caption{VEIL Pipeline. We convert time series into chart-based visual encodings (line, area, bar, scatter), learn features using CNNs, analyze encoding sensitivity through representation, transfer, and attribution methods, and apply 
   HINT-based refinement to guide model attention.}

  \label{fig:system}
  \vspace{-1.2em}
\end{figure*}
\paragraph{Image-Based Time-Series Classification.}

Encoding numerical time series as images for CNN-based classification has been explored through a range of representations, including Gramian Angular Fields and Markov Transition Fields~\cite{wang2015imaging}, Recurrence Plots and their fusions~\cite{hatami2018classification, mariani2024fusion}, and time–frequency transforms such as CWT scalegrams~\cite{zhao2024cwt3dcnn, alghanemi2024cwt}. These approaches are grounded in the idea of spatially encoding temporal correlations to make sequential data amenable to vision models \cite{venkatesan2026vtbenchmultimodalframeworktimeseries}. More recent work, summarized in surveys~\cite{ni2025harnessing}, reflects a broader shift toward leveraging powerful vision backbones, including pretrained transformers, for time series analysis, alongside chart-style renderings like line plots~\cite{li2023vitst} and screenshot-based encodings~\cite{oh2025tssi} rooted in human visualization practices. Prior work typically commits to a single encoding and evaluates predictive performance under that design, demonstrating effectiveness but giving less attention to systematically isolating the role of encoding itself or understanding whether performance arises from temporal structure or encoding-specific visual patterns. Our work builds on this line of research by explicitly analyzing how different chart encodings influence learned representations and model behavior.
Rendering one-dimensional signals as two-dimensional charts lets TSC models leverage pretrained vision backbones trained on diverse image corpora. This transformation also makes large-scale numerical streams more accessible for human inspection, since tools such as Grad-CAM can be overlaid on chart images to support visual verification of model decision logic. Visual encodings may further introduce redundant geometric structure that helps CNNs capture shape-level patterns, though this benefit depends on whether the rendering preserves task-relevant signal properties \cite{islam2021shape}.


\paragraph{Visual Feature Bias and Representational Analysis.}
Deep neural networks often exploit dataset-specific shortcuts over robust semantics ~\cite{geirhos2019texture, geirhos2020shortcut}. In vision-based pipelines, this manifests as reliance on visually salient but weak patterns ~\cite{suhail2025shortcut}; in chart-based TSC, such shortcuts may reflect encoding artifacts rather than true temporal dynamics. Prior work offers complementary diagnostics: representation similarity methods such as Centered Kernel Alignment (CKA) ~\cite{kornblith2019similarity} compares feature spaces and has assessed how models perceive visualizations ~\cite{long2025seeing}; linear probing ~\cite{alain2017probing} tests generality; gradient-based attribution ~\cite{selvaraju2017gradcam} reveals model focus; and HINT (Human Importance-aware Network Tuning) ~\cite{selvaraju2019hint} redirects attention to semantic regions, though unexplored in chart-based TSC. Visualization research further shows that chart design shapes perception of temporal patterns ~\cite{proma2025evaluating}. We use a representative subset of these tools to test whether chart-based TSC models learn temporal structure or rely on encoding-driven visual artifacts. 
\section{Methodology}
\label{sec:method}

\subsection{Experimental Setup}
\label{subsec:setup}

We evaluate four standard chart types—line, area, bar, and scatter—as visual encodings for TSC (Fig.~\ref{fig:system}A). Each series is rendered at $128{\times}128$ pixels using matplotlib~\cite{hunter2007matplotlib}, preserving key visual features (e.g., slopes, bar heights, point density). To isolate the effect of encoding, all models share an identical training protocol on standard UCR splits~\cite{middlehurst2024bakeoff}: a CNN backbone (Fig.~\ref{fig:system}B), an Adam optimizer~\cite{kingma2015adam} with default hyperparameters ($\text{lr}{=}10^{-3}$, weight decay${=}10^{-2}$), and early stopping (patience $10$) \cite{venkatesan2026vtbenchmultimodalframeworktimeseries}. No per-dataset tuning is performed, ensuring that performance differences arise from encoding rather than optimization choices.

\subsection{Analysis Methods}
\label{subsec:diagnostics}
To understand how chart encodings influence learned representations and model behavior, we employ complementary analysis methods spanning three categories: representation similarity, transferability, and attribution, as shown in Fig. \ref{fig:system}C.

\vspace{0.5em}
\noindent
\textbf{Representational Similarity (CKA).} We measure alignment between feature representations learned from different chart types using linear Centered Kernel Alignment (CKA)~\cite{kornblith2019similarity}. Let $\mathbf{H}_a \in \mathbb{R}^{n \times d_a}$ and $\mathbf{H}_b \in \mathbb{R}^{n \times d_b}$ denote the activation matrices for two models $a$ and $b$ over the same $n$ test samples,
\begin{equation}
  \mathrm{CKA}(X, Y) = \frac{\|Y^\top X\|_F^2}{\|X^\top X\|_F \, \|Y^\top Y\|_F},
\end{equation}
which is invariant to orthogonal transformations and isotropic
scaling and is well-suited for comparing representations across
encoders trained on visually distinct inputs.

\vspace{0.5em}
\noindent
\textbf{Cross-Chart Linear Probing.} To evaluate transferability, we freeze each encoder and train a linear classifier (we have chosen a simple logistic regression model \cite{chung2020introduction}) on features extracted from a \emph{different} chart type. Cross-chart probe accuracy provides a lower bound on shared task-relevant information and complements CKA by measuring alignment in label space.

\vspace{0.5em}
\noindent
\textbf{Intrinsic Dimensionality.}
We implement PCA~\cite{jolliffe2002principal}, a dimensionality reduction technique used to simplify complex datasets while retaining maximum variance. This is applied to the penultimate features of each type of chart encoding, reporting the smallest number of components required to explain 90\% of variance, as a coarse measure of representation compactness.

\vspace{0.5em}
\noindent
\textbf{Feature Space Visualization.}
We visualize feature geometry using UMAP~\cite{mcinnes2018umap}, which projects high-dimensional representations into two dimensions while preserving local structure. This enables qualitative inspection of class separation and clustering patterns across encodings.

\vspace{0.5em}
\noindent
\textbf{Sensitivity Indices}
To generalize our findings beyond the initial exploratory analysis, we scale the CKA and cross-probe transfer experiments to the 31 datasets from the UCR benchmark. For each dataset, we compute a CKA similarity matrix and a cross-chart probe transfer matrix, then derive scalar metrics (Encoding Sensitivity Indices):  $\text{ESI}_{\text{CKA}} = 1 - \text{mean off-diagonal CKA}$ and $\text{ESI}_{\text{Probe}} = 1 - \text{mean off-diagonal probe accuracy}$. Higher values indicate greater sensitivity to encoding choices. These indices provide a dataset-level view of encoding invariance versus dependence.

\vspace{0.5em}
\noindent
\textbf{Grad-CAM Attribution.}
\begin{figure}[t]
  \centering
  \begin{minipage}{0.32\linewidth}
    \centering
    \includegraphics[width=\linewidth]{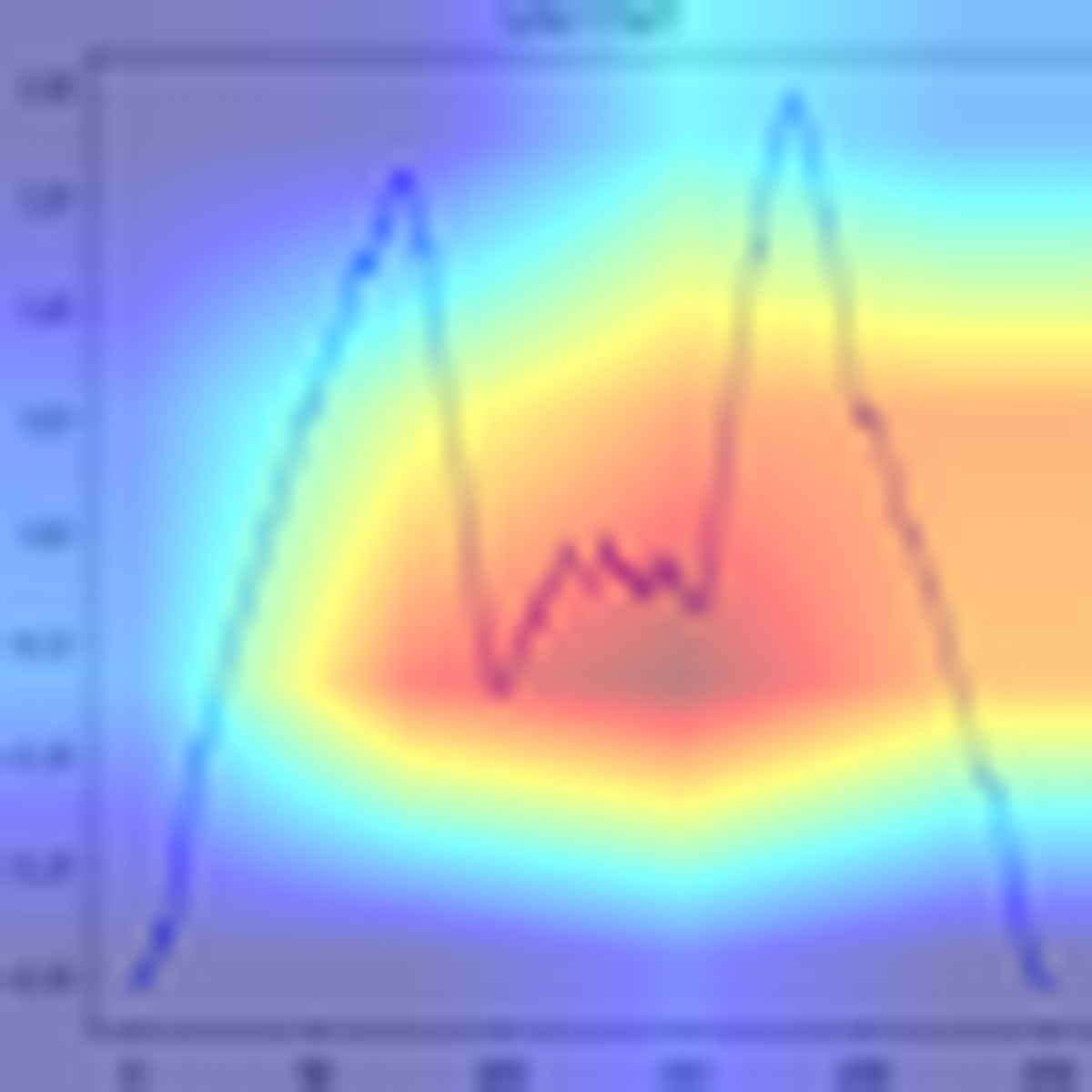}
    \subcaption{ArrowHead - line}
  \end{minipage}\hfill
  \begin{minipage}{0.32\linewidth}
    \centering
    \includegraphics[width=\linewidth]{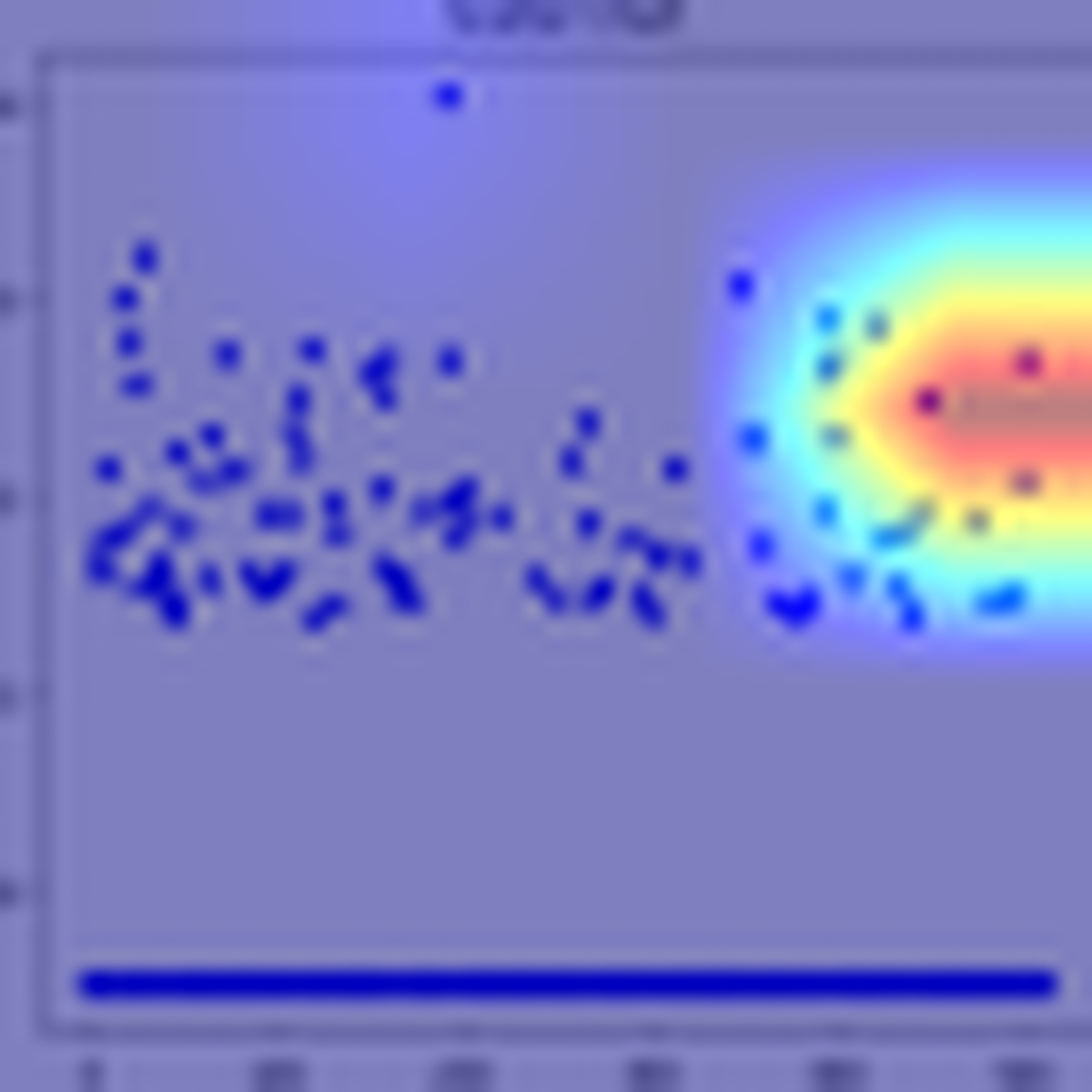}
    \subcaption{Earthquakes - scatter}
  \end{minipage}\hfill
  \begin{minipage}{0.32\linewidth}
    \centering
    \includegraphics[width=\linewidth]{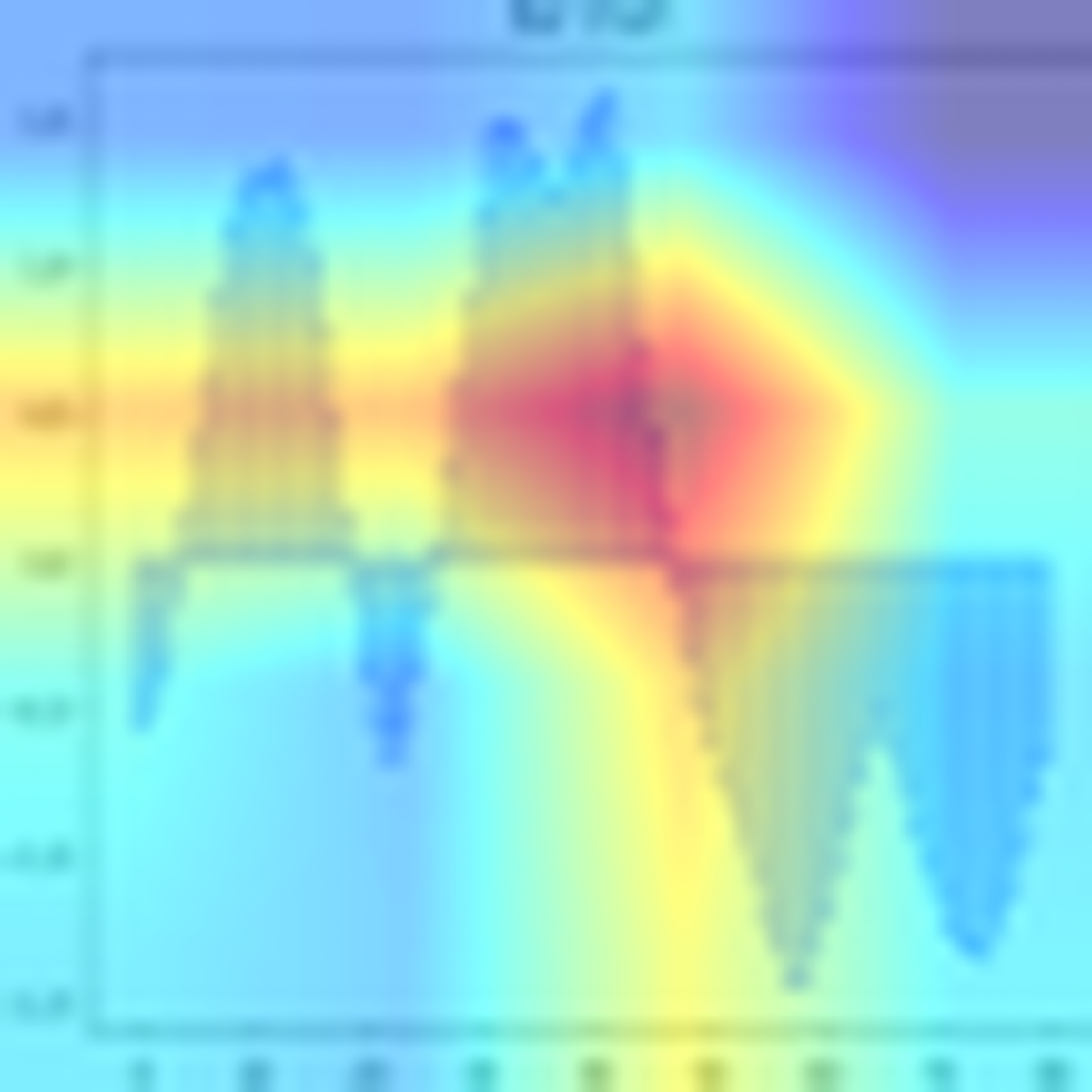}
    \subcaption{Phalanges - bar}
  \end{minipage}
  \caption{Grad-CAM heatmap overlays on original chart images for ArrowHead (line), Earthquakes (scatter) and PhalangesOutlinesCorrect (bar). The color intensity indicates regions where the model focuses most when making its prediction, with warmer colors representing higher importance.}
  \label{fig:gradcam_overlays}
  \vspace{-1.8em}
\end{figure}
To examine model attention, we apply Gradient-weighted Class Activation Mapping (Grad-CAM) \cite{selvaraju2017gradcam} to the final convolutional layer of each trained CNN (Fig.~\ref{fig:gradcam_overlays}).  Grad-CAM produces class-specific saliency maps, allowing us to distinguish whether models rely on (i) signal structure (e.g., temporal trends), (ii) encoding-specific features (e.g., bar edges, scatter density), or (iii) irrelevant elements (e.g., axes, gridlines). We analyze 10 correctly and 10 incorrectly classified samples per class and chart type, overlaying heatmaps on input images to compare attention patterns across encoding regimes.

\vspace{0.5em}
\noindent
\textbf{Chart Perturbations.}
To stress-test model sensitivity to rendering-dependent visual evidence, we apply controlled perturbations as diagnostic probes: Gaussian blur (line/area charts), bar merging (bar charts), and alpha fading (scatter plots). Performance degradation ($\Delta$ accuracy) therefore flags encoding sensitivity, but does not by itself establish that the disrupted evidence is task-irrelevant.


\vspace{0.5em}
\noindent
\textbf{Generalization Across Encoding Families.}
As a robustness check, we evaluate whether the hijacking effect persists beyond chart-style rendering, we additionally evaluate five mathematical transforms (GASF, MTF~\cite{wang2015imaging}, RP~\cite{eckmann1987recurrence}, CWT~\cite{mallat1989wavelet}, STFT~\cite{griffin1984stft}) under the same training protocol. This tests whether observed trends extend beyond chart-style representations.

\subsection{HINT Attention Guidance}
\label{subsec:hint}
To test whether model attention can be redirected toward semantically meaningful temporal regions, we apply HINT-based attention guidance (Figure \ref{fig:system}D). Before computing the alignment loss, we mask the most salient Grad-CAM regions with a white bounding box, occluding them to isolate the model's reliance on those cues. The masked image is then passed through the model to produce a second Grad-CAM map, used as the alignment signal. The training objective combines classification and attention alignment losses: $\mathcal{L}_{\text{total}} = \mathcal{L}_{\text{cls}} + \lambda \cdot \mathcal{L}_{\text{HINT}}$, where $\mathcal{L}_{\text{HINT}}$ penalizes divergence between the model's Grad-CAM map and a target attention mask derived from temporal importance annotations. Unlike prior work relying on human annotations, our approach derives supervision from model-driven saliency, enabling a self-supervised adaptation of HINT for chart-based TSC.


\section{Results}
\label{sec:results}

\subsection{Representational Divergence Across Chart Types}
\label{subsec:cka_results}

\begin{figure}[t!]
  \centering
  \begin{minipage}{0.48\linewidth}
    \centering
    \includegraphics[width=\linewidth]{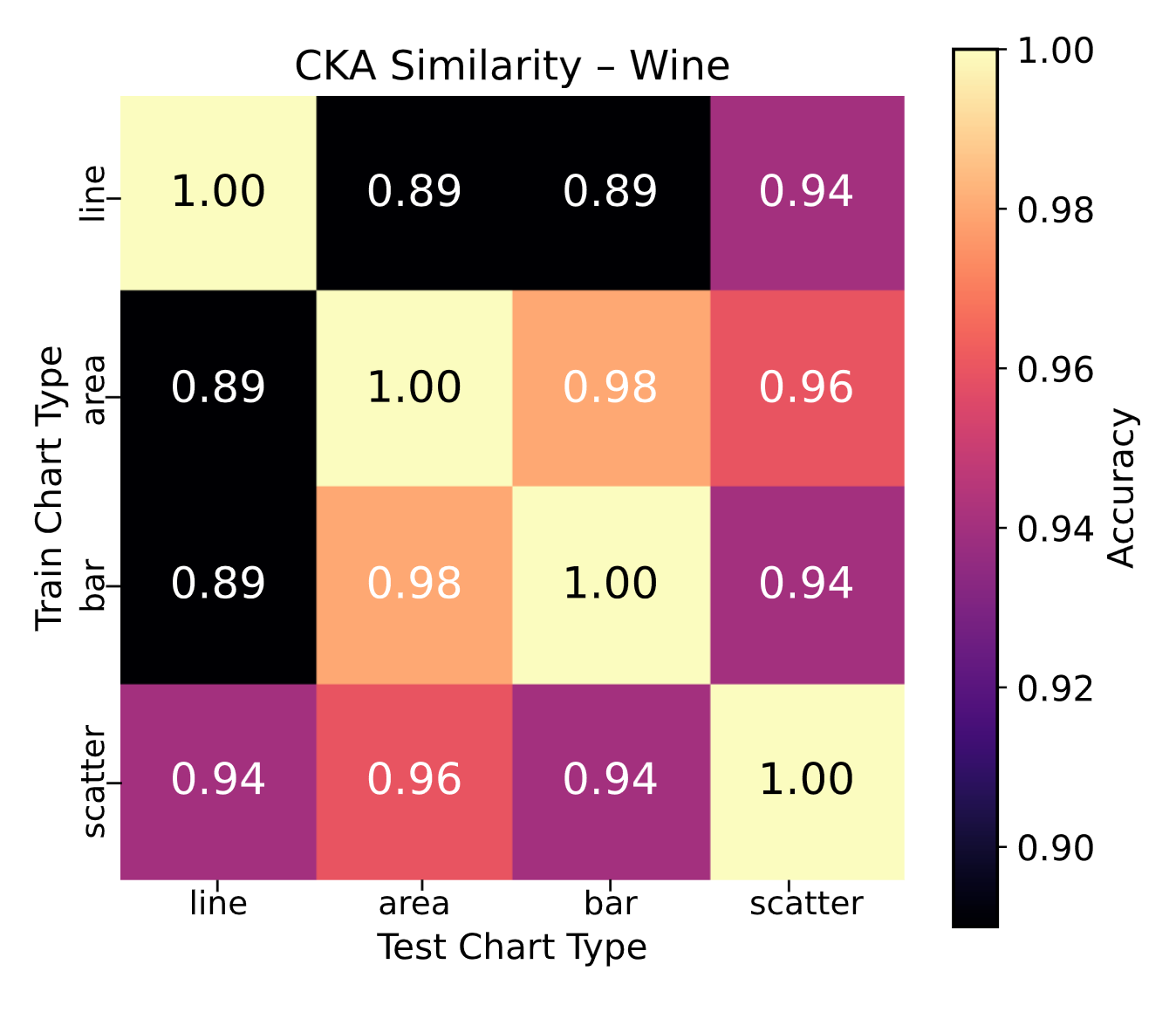}
    \subcaption{Wine}
  \end{minipage}\hfill
  \begin{minipage}{0.47\linewidth}
    \centering
    \includegraphics[width=\linewidth]{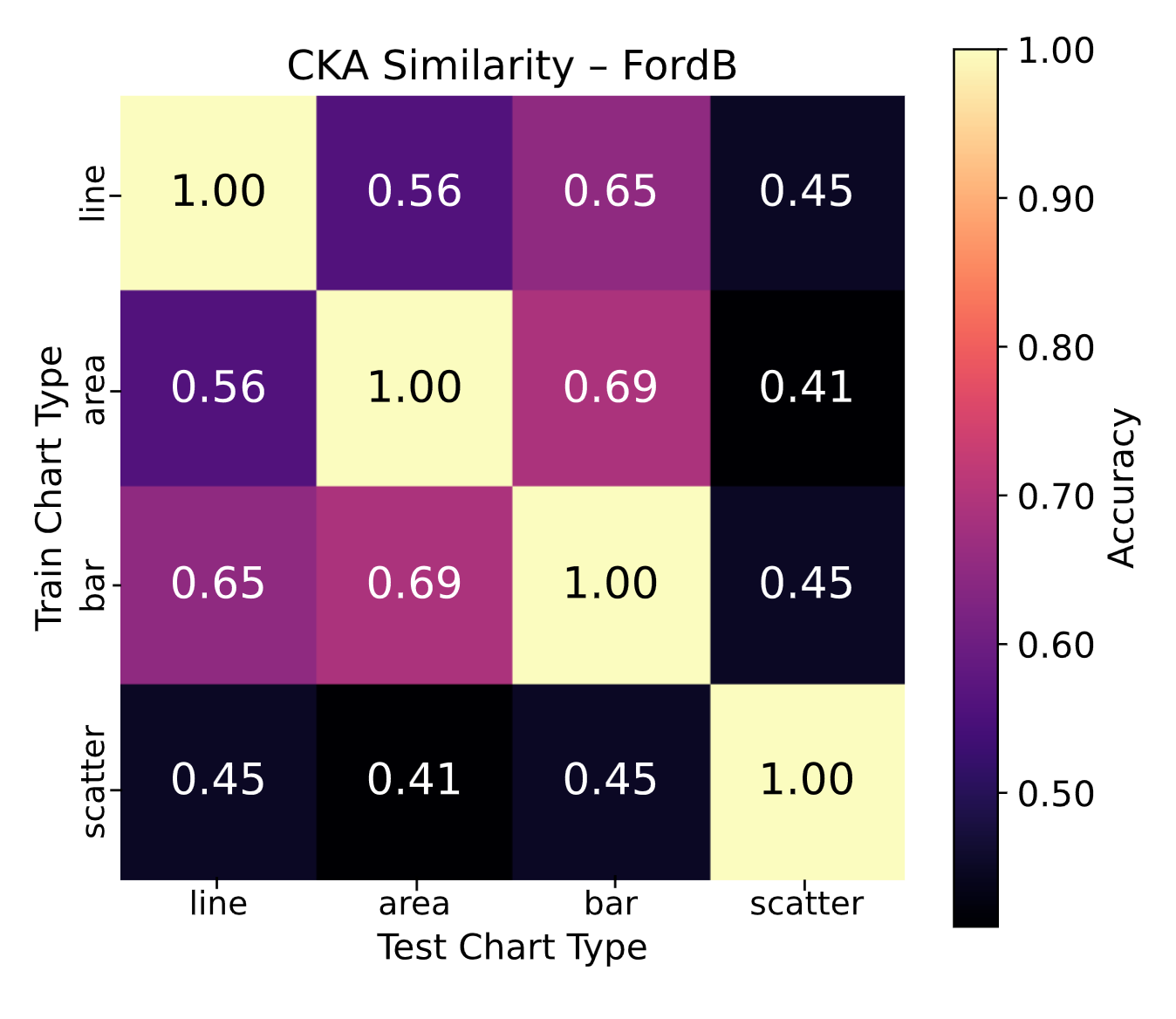}
    \subcaption{FordB}
  \end{minipage}\\[0.1cm]
  \begin{minipage}{0.48\linewidth}  
    \centering
    \includegraphics[width=\linewidth]{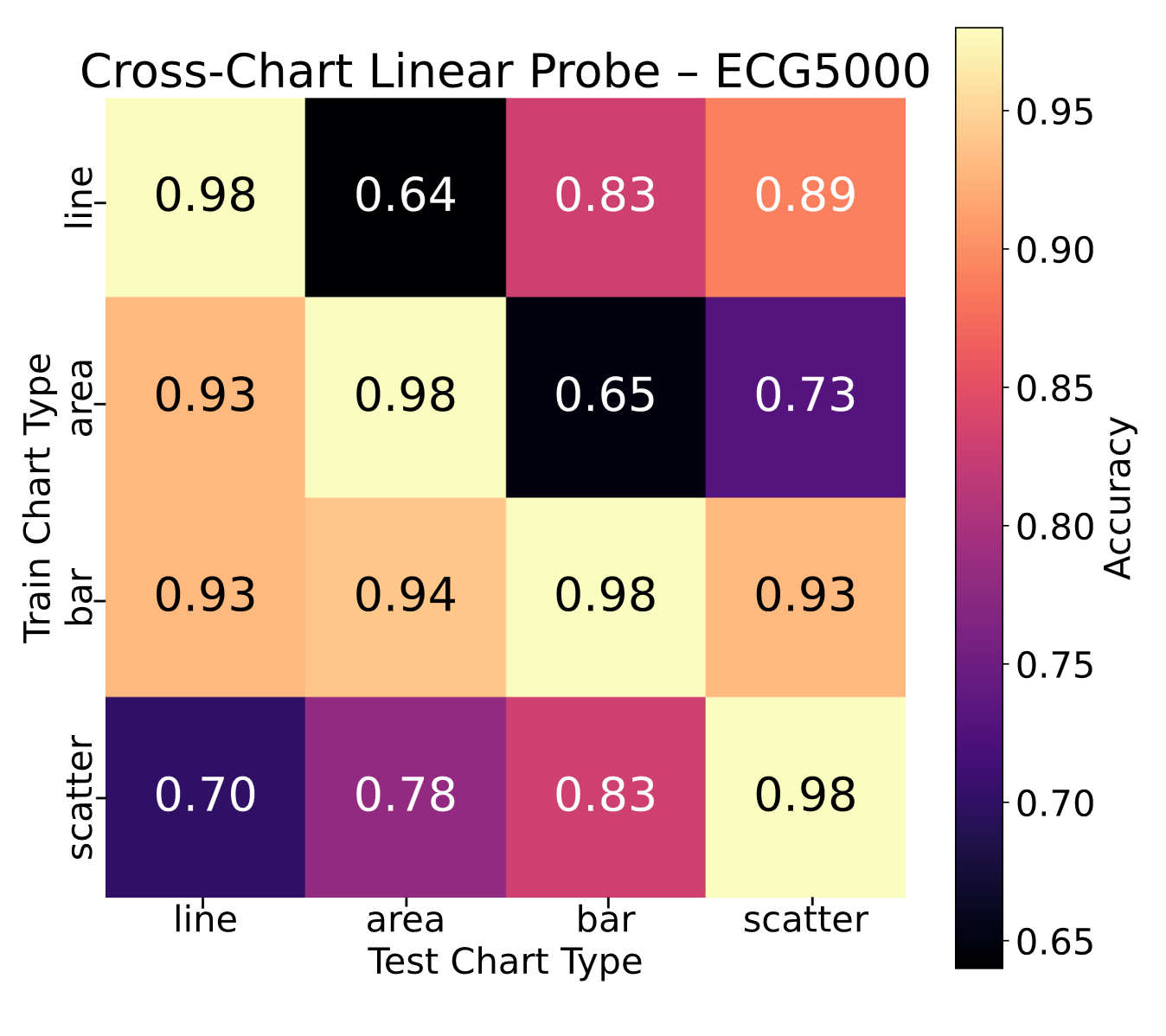}
    \subcaption{ECG5000}
  \end{minipage}\hfill
  \begin{minipage}{0.48\linewidth}  
    \centering
    \includegraphics[width=\linewidth]{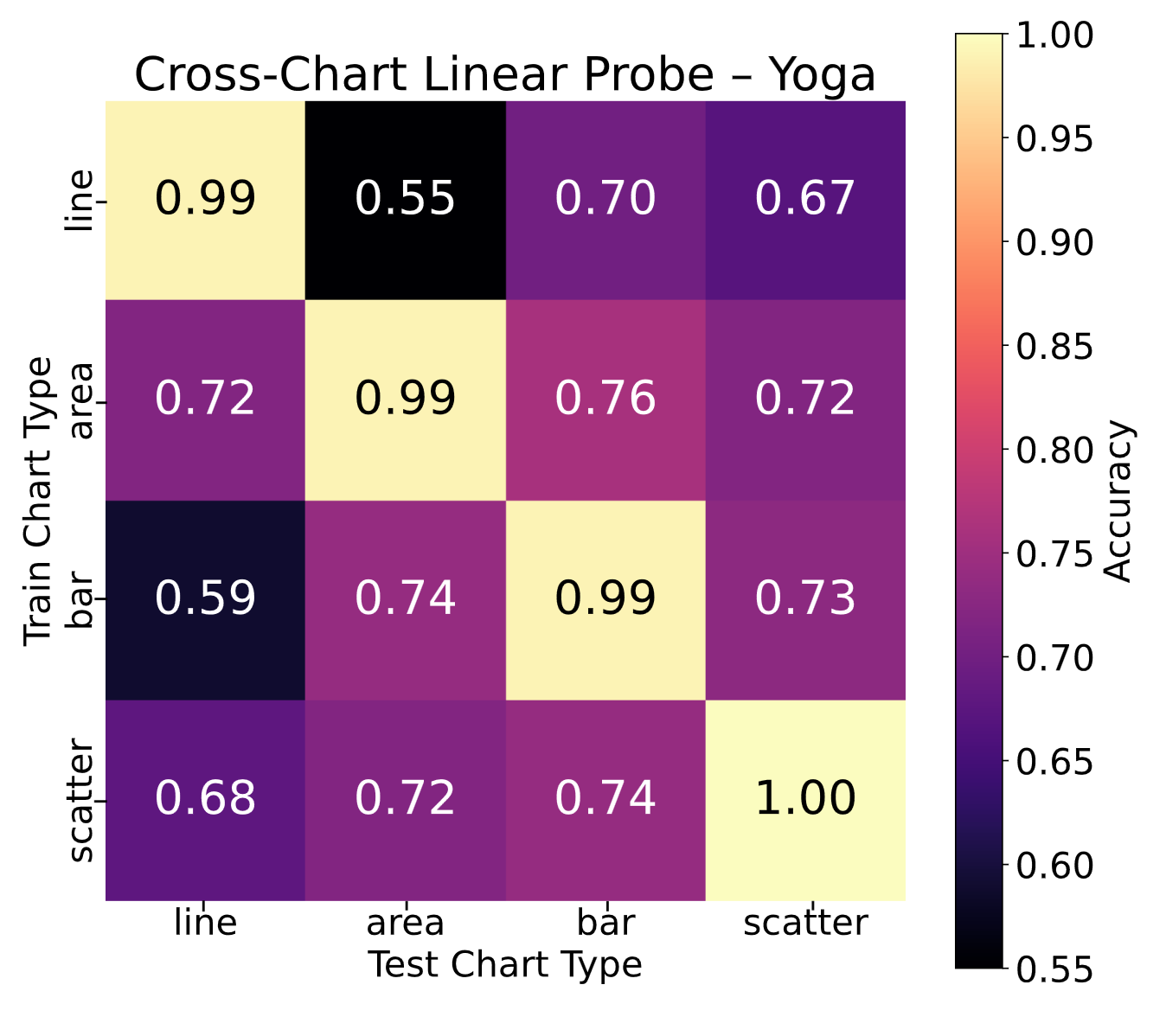}
    \subcaption{Yoga}
  \end{minipage}
  \caption{(a)--(b) CKA similarity matrices across chart types on Wine (encoding-invariant) and FordB (encoding-divergent). (c)--(d) Cross-chart linear probe accuracy on ECG5000 (high transfer) and Yoga (low transfer); off-diagonals quantify encoding invariance.}
  \label{fig:cka_linear_probe}
  \vspace{-2em}
\end{figure}
\textbf{Encoding alignment is bimodal; representations from simple datasets achieve high convergence, whereas those from complex datasets show substantial divergence (RQ1, RQ2).} CKA analysis reveals a relationship between dataset complexity and representational alignment across chart types. As seen in Fig.~\ref{fig:cka_linear_probe}(a), simple binary-class datasets (Wine) show high CKA scores ($\approx$0.89--0.98), indicating encoding-invariant representations. Complex, multi-class datasets like FordB (Fig.~\ref{fig:cka_linear_probe}(b)) exhibit minimal cross-chart similarity ($\approx$0.44--0.69), suggesting that chart type may fundamentally alter the learned feature space. Moderate-complexity datasets like GunPoint and Yoga fall in between, with area-bar pairs showing strong alignment ($\approx$0.86--0.96) while line and scatter diverge ($\approx$0.67--0.79). Cross-chart transferability follows this pattern (Fig.~\ref{fig:cka_linear_probe}(c-d)): encoding-invariant datasets like ECG5000 exhibit high off-diagonal accuracy (0.64–0.94), indicating that features learned from one chart type generalize effectively. In contrast, encoding-sensitive datasets like Yoga show a gap between diagonal (0.99–1.00) and off-diagonal (0.55–0.76) accuracy, suggesting that models may specialize in the visual patterns of their training encoding with limited transfer to unseen chart types.

\textbf{Chart-type separation persists across dataset lengths; encoding shapes geometry independently of intrinsic dimensionality (RQ1).}
\begin{figure}[t!]
  \centering
  \begin{minipage}{0.83\linewidth}
    \centering
    \includegraphics[width=\linewidth]{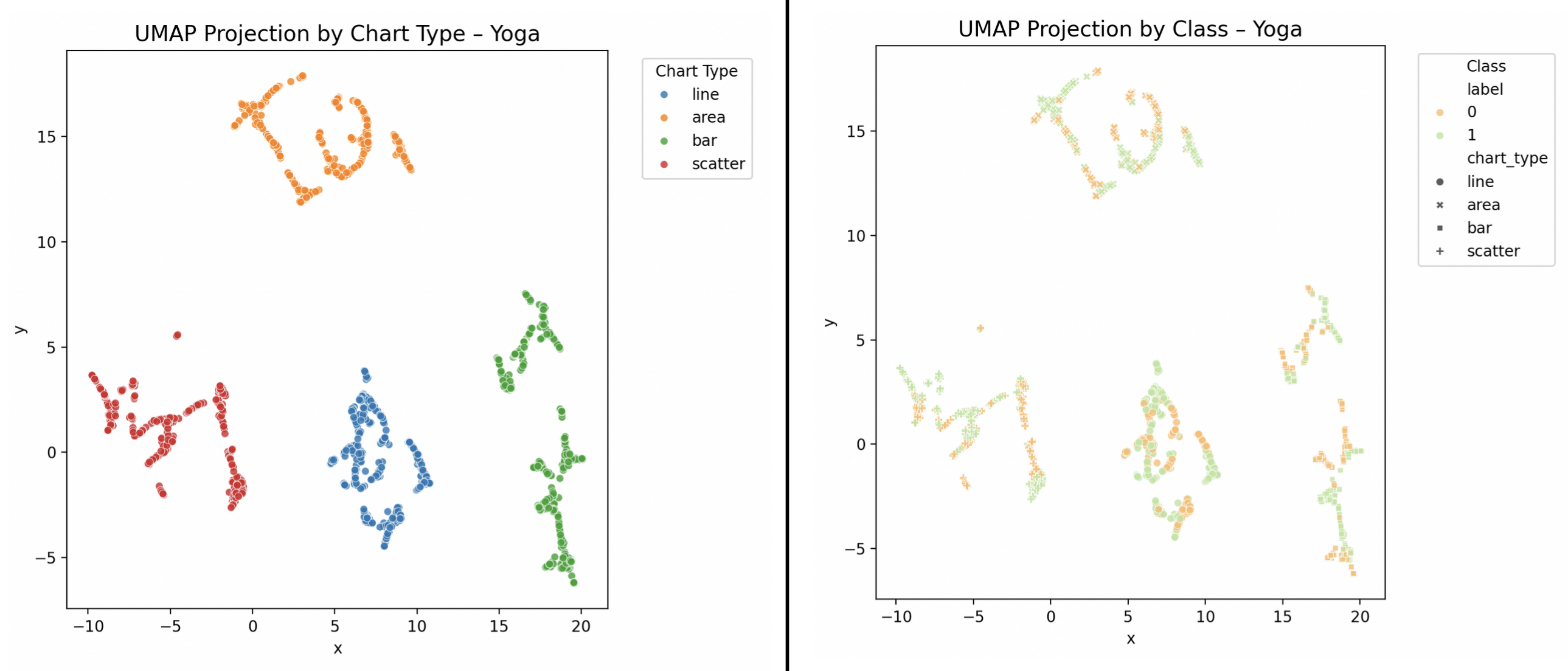}
    \subcaption{UMAP- Yoga}
  \end{minipage}\hfill
  \begin{minipage}{0.83\linewidth}
    \centering
    \includegraphics[width=\linewidth]{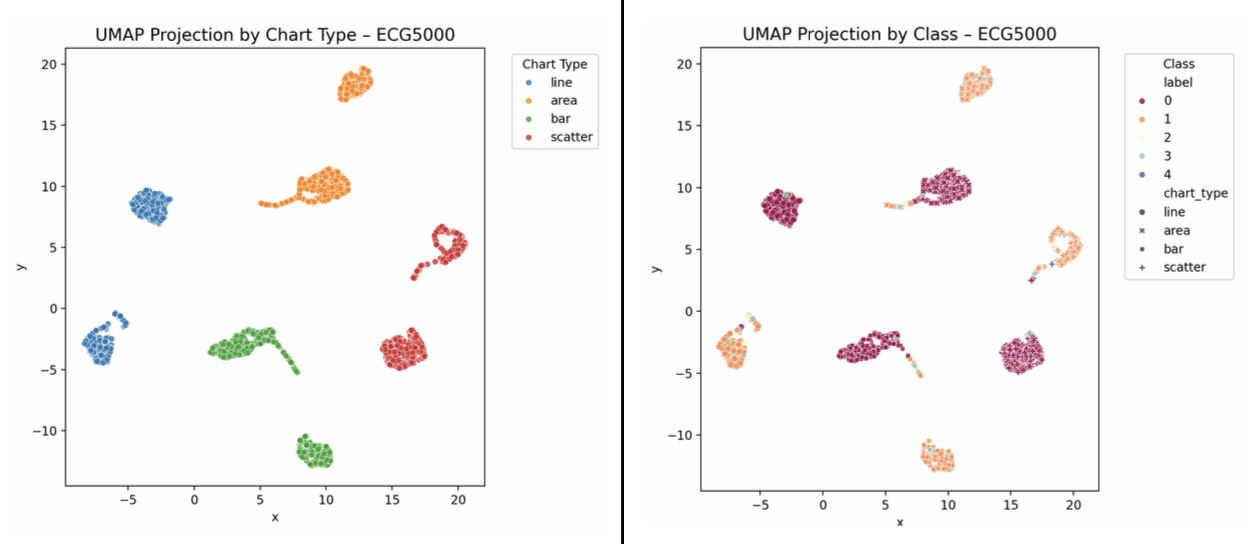}
    \subcaption{UMAP- ECG5000}
  \end{minipage}
  \caption{(a)-(b) UMAP projections of learned embeddings colored by chart type and class label for FordB and ECG5000, revealing whether models cluster by visual encoding or semantic class.}
  \label{fig:pca_umap}
  \vspace{-1.8em}
\end{figure}
PCA and UMAP together reveal how dataset length \cite{venkatesan2026vtbenchmultimodalframeworktimeseries} governs representational structure. Short datasets (e.g., ECG5000, GunPoint) exhibit low intrinsic dimensionality (1–11 components), and Fig. \ref{fig:pca_umap}(b) shows clear UMAP separation by both chart type and class. Medium datasets (e.g., Strawberry, FordB) remain low-dimensional (1–2 components) but show stronger chart-type clustering, implying increased reliance on visual encoding. Long datasets like Yoga exhibit higher dimensionality (22–26 components) with diffuse class boundaries yet clear chart-type clustering (Fig.~\ref{fig:pca_umap}(a)), indicating that extended temporal structure expands representational capacity while amplifying encoding-specific sensitivity. Chart type minimally impacts dimensionality, confirming that dataset complexity, not visualization form, drives variance.

\begin{figure}[!t]
    \centering
    \includegraphics[width=0.48\textwidth]{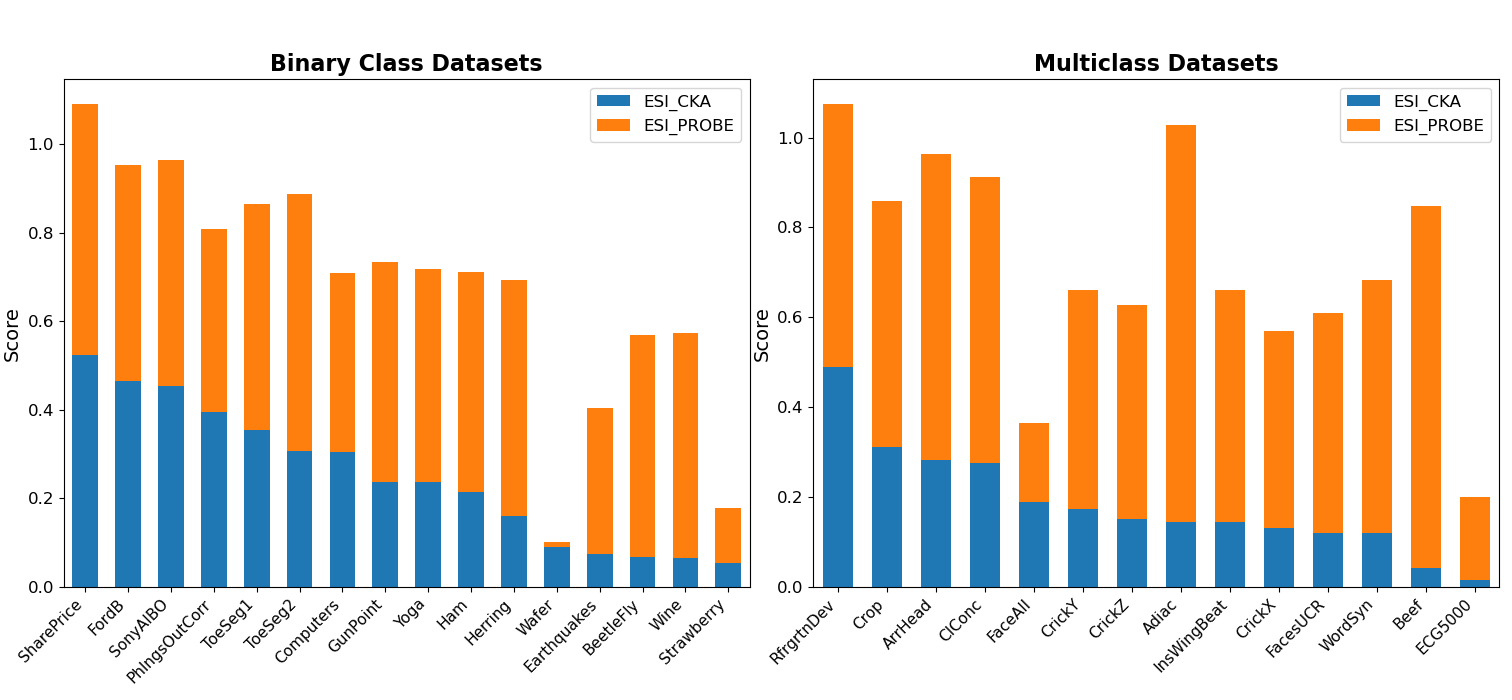}
    \caption{Encoding Sensitivity Index (ESI) values for CKA and linear probe across all datasets, separated by binary class datasets and multi-class datasets.}
    \label{fig:esi}
\end{figure}

\textbf{Encoding sensitivity is more strongly associated with class count than sequence length in our dataset-level analysis (RQ1, RQ2).} 
As shown in Fig.~\ref{fig:esi}, $\text{ESI}_{\text{CKA}}$ peaks for SharePriceIncrease (0.524), FordB (0.464), and RefrigerationDevices (0.49), while $\text{ESI}_{\text{Probe}}$ peaks for Adiac (0.883) and Beef (0.806). Cross-metric dissociations also emerge: Beef shows low representational alignment but high probe sensitivity, whereas Wafer shows relatively stable representations but weaker linear transfer. Notably, binary datasets such as Wafer and Strawberry exhibit low encoding sensitivity despite differences in sequence length, while multi-class datasets such as Adiac show elevated sensitivity even with shorter sequences~\cite{venkatesan2026vtbenchmultimodalframeworktimeseries}. These patterns suggest that class structure may be a stronger correlate of encoding sensitivity than sequence length, but further controlled analysis is needed before making a causal claim.

\begin{figure}[!t]
  \centering
  \includegraphics[width=\linewidth]{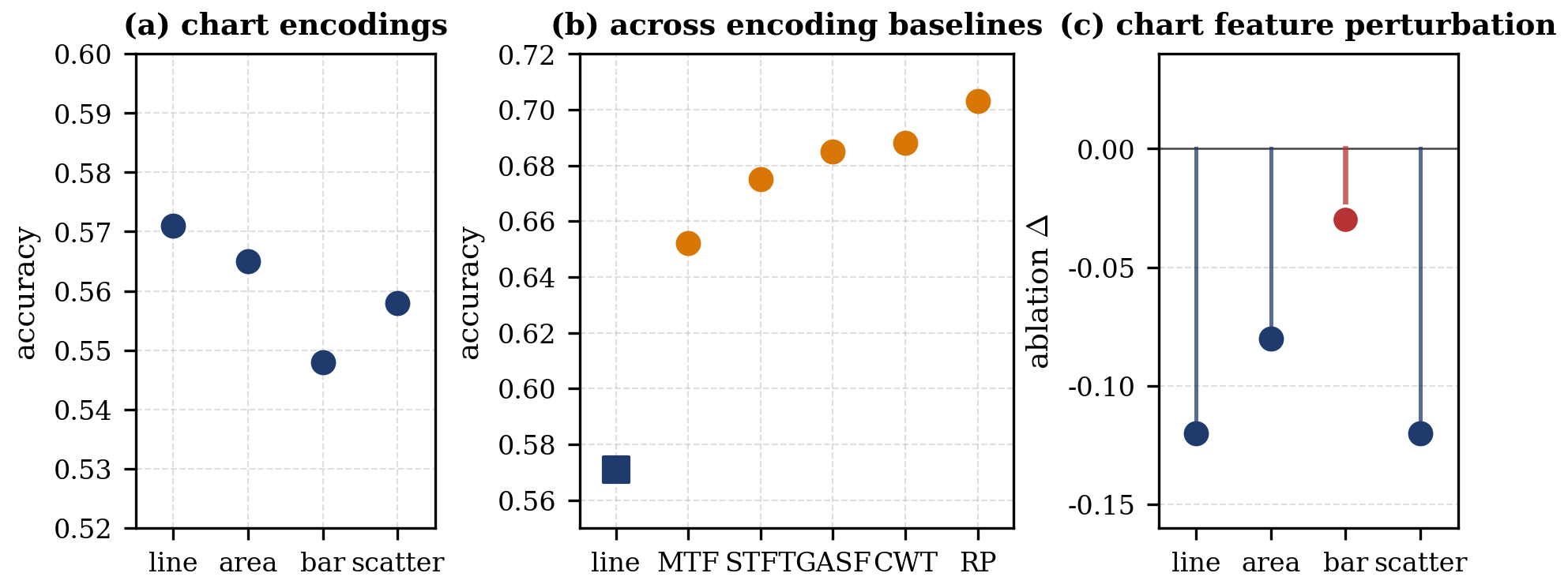}
  \caption{Analysis of impact of chart-feature perturbations. (a) Four encodings yield distinct accuracies (max--min $\Delta = 0.023$ across 31 datasets). (b) Comparison with cross-encoding baselines highlight differences in representational effectiveness. (c) Perturbation degrades accuracy unevenly across encodings; smaller drops for bar charts ($-0.03$) suggest lower reliance on fine-grained visual cues, consistent with encoding-dependent behavior.}
  \label{fig:ablation}
  \vspace{-1.9em}
  \end{figure}

\subsection{Model Behavior Under Controlled Perturbations}
\textbf{Model behavior is consistently associated with encoding-dependent inductive biases rather than encoding-invariant representations (RQ3).} In VEIL, no encoding consistently dominates across datasets, suggesting that chart features act as distinct inductive priors rather than interchangeable inputs (Fig.~\ref{fig:ablation}a). Performance differences become more pronounced across encodings, indicating that encoding choice can substantially alter the feature distributions learned by the model (Fig.~\ref{fig:ablation}b). Targeted perturbations further serve as diagnostic stress tests, revealing sensitivity to rendering-dependent visual evidence: performance drops when key visual elements are disrupted, while bar charts remain comparatively robust, suggesting lower sensitivity to fine-grained features and greater reliance on coarse structural patterns (Fig.~\ref{fig:ablation}c). These results show sensitivity to encoding-specific visual elements, but do not establish that the disrupted evidence was task-irrelevant.

\subsection{HINT Attention Guidance}

\begin{table}[t!]
  \tiny 
  \centering
  \caption{HINT implementation results: best-performing encoding per dataset, with baseline accuracy, HINT accuracy, and delta value of improvement. Positive deltas indicate improvement from attention guidance; negative deltas indicate degradation.}
  \label{tab:hint_results}
  \begin{tabular}{lccccc}
    \toprule
    \textbf{Dataset} & \textbf{Best Encoding} & \textbf{Baseline (\%)} & \textbf{HINT (\%)} & \textbf{Delta (\%)} \\
    \midrule
    \rowcolor{green!15} ArrowHead & Area & 30.29 & 39.43 & \textbf{+9.14} \\
    \rowcolor{red!15} PhalangesOutlinesCorrect & Scatter & 65.15 & 61.31 & -3.85 \\
    \rowcolor{red!15} ChlorineConcentration & Bar & 55.78 & 53.26 & -2.53 \\
    \rowcolor{red!15} SonyAIBORobotSurface1 & Line & 57.07 & 42.93 & -14.14 \\
    \rowcolor{green!15} Adiac & Scatter & 22.51 & 39.13 & \textbf{+16.62} \\
    \rowcolor{red!15} FaceAll & Scatter & 77.93 & 75.92 & -2.01 \\
    \rowcolor{green!15} FacesUCR & Area & 57.12 & 77.66 & \textbf{+20.54} \\
    \rowcolor{green!15} CricketX & Scatter & 52.56 & 63.59 & \textbf{+11.03} \\
    \rowcolor{green!15} CricketY & Scatter & 55.38 & 64.10 & \textbf{+8.72} \\
    \rowcolor{green!15} CricketZ & Area & 54.36 & 65.38 & \textbf{+11.03} \\
    \rowcolor{green!15} ToeSegmentation1 & Scatter & 47.37 & 55.70 & \textbf{+8.33} \\
    \rowcolor{green!15} ToeSegmentation2 & Area & 43.08 & 81.54 & \textbf{+38.46} \\
    Wine & Bar & 50.00 & 50.00 & 0.00 \\
    \rowcolor{green!15} InsectWingbeat & Area & 57.24 & 63.68 & \textbf{+6.34} \\
    \rowcolor{green!15} WordSynonyms & Scatter & 48.75 & 53.76 & \textbf{+5.02} \\
    Beef & Scatter & 20.00 & 20.00 & 0.00 \\
    BeetleFly & Scatter & 50.00 & 50.00 & 0.00 \\
    \rowcolor{red!15} Computers & Bar & 69.60 & 33.60 & -36.00 \\
    \rowcolor{green!15} Earthquakes & Scatter & 69.06 & 74.82 & \textbf{+5.76} \\
    \rowcolor{green!15} Ham & Area & 50.48 & 74.29 & \textbf{+23.81} \\
    \rowcolor{green!15} Herring & Area & 40.62 & 59.38 & \textbf{+18.75} \\
    \rowcolor{red!15} RefrigerationDevices & Area & 54.13 & 53.87 & -0.27 \\
    \rowcolor{green!15} SharePriceIncrease & Line & 58.59 & 68.63 & \textbf{+10.04} \\
    \rowcolor{green!15} Crop & Bar & 47.68 & 68.60 & \textbf{+20.92} \\
    \rowcolor{red!15} ECG5000 & Line & 93.44 & 93.29 & -0.16 \\
    \rowcolor{green!15} GunPoint & Area & 49.33 & 61.33 & \textbf{+12.00} \\
    \rowcolor{green!15} Strawberry & Bar & 90.00 & 93.78 & \textbf{+3.78} \\
    \rowcolor{green!15} FordB & Bar & 62.96 & 73.58 & \textbf{+10.62} \\
    \rowcolor{green!15} Wafer & Area & 99.50 & 99.59 & \textbf{+0.10} \\
    \rowcolor{green!15} Yoga & Line & 49.80 & 52.60 & \textbf{+2.80} \\
    Lightning2 & Area & 54.10 & 54.10 & 0.00 \\
    \bottomrule
  \end{tabular}
  \vspace{-2.5em}
\end{table}
\textbf{HINT improves performance on several encoding-divergent datasets but is not a general mitigation strategy (RQ3).}
As shown in Table \ref{tab:hint_results}, HINT gains align with prior diagnostics. Encoding-divergent datasets (low CKA, low cross-chart transfer, high $\text{ESI}_{\text{Probe}}$ $> 0.5$, UMAP clustering by chart type) show substantial improvements: ToeSegmentation2 (+38.46\%), Ham (+23.81\%), Crop (+20.92\%), and FacesUCR (+20.54\%). Encoding-invariant datasets (high CKA, strong transfer, low $\text{ESI}_{\text{Probe}}$ $< 0.2$, UMAP mixing by class) show minimal effects (ECG5000: -0.16\%, Wafer: +0.10\%). PCA confirms that dimensionality alone does not predict HINT effectiveness. Area and scatter encodings benefited most, consistent with UMAP evidence of distinct clustering. However, HINT degraded performance for Computers (-36.00\%), SonyAIBORobotSurface1 (-14.14\%), and PhalangesOutlinesCorrect (-3.85\%), suggesting that attention guidance may interfere with already-effective representations. These results suggest that attention guidance can reduce encoding-sensitive behavior in selected cases when supported by convergent evidence across CKA, linear probe, ESI, UMAP, and PCA, but its failures indicate that HINT should be treated as a diagnostic intervention rather than a reliable mitigation method.




\section{Discussion and Conclusion}
\label{sec:conclusion}
Our results suggest a fundamental tension in image-based TSC: models appear to learn both temporal structure and encoding-specific visual cues. We term this \emph{visual encoding hijacking}, associated with cases where performance may depend on how a signal is rendered, not what it represents. \textbf{(1) Encoding as inductive bias:} visual encoding appears to be a potential source of inductive bias, as chart types produce divergent feature spaces and inconsistent transfer, so encoding choice may warrant the same care as model design. \textbf{(2) Rethinking performance gains:} high accuracy does not necessarily imply semantic understanding, since models may exploit encoding-specific artifacts, raising concerns about cross-encoding comparisons. \textbf{(3) Toward diagnostic evaluation:} accuracy alone is insufficient; combining similarity, transferability, attribution, and perturbation stress tests reveals encoding sensitivity. Perturbations do not prove that encoding-dependent evidence is task-irrelevant. Overall, VEIL addresses a growing question: how do vision models perceive charts? While graphical perception studies human interpretation of visualizations, VEIL examines model interpretation, showing that visual encoding choices shape learned representations, though these effects do not always reflect semantic understanding. Encoding choice substantially affects what models learn, making chart design consequential for both human communication and machine consumption. These findings motivate encoding-aware evaluation protocols that better align model reasoning with human-interpretable structure.

\acknowledgments{
This work was supported in part by the U.S. National Science Foundation under Grant No. IIS-2427770.}

\bibliographystyle{abbrv-doi}
\bibliography{main}

@article{wu2022timesnet,
  title={Timesnet: Temporal 2d-variation modeling for general time series analysis},
  author={Wu, Haixu and Hu, Tengge and Liu, Yong and Zhou, Hang and Wang, Jianmin and Long, Mingsheng},
  journal={arXiv preprint arXiv:2210.02186},
  year={2022}
}

@inproceedings{lee2024assessing,
  title={Assessing graphical perception of image embedding models using channel effectiveness},
  author={Lee, Soohyun and Chang, Minsuk and Park, Seokhyeon and Seo, Jinwook},
  booktitle={2024 IEEE Visualization and Visual Analytics (VIS)},
  pages={226--230},
  year={2024},
  organization={IEEE}
}

@article{ni2025harnessing,
  title={Harnessing vision models for time series analysis: A survey},
  author={Ni, Jingchao and Zhao, Ziming and Shen, ChengAo and Tong, Hanghang and Song, Dongjin and Cheng, Wei and Luo, Dongsheng and Chen, Haifeng},
  journal={arXiv preprint arXiv:2502.08869},
  year={2025}
}

@article{suhail2025shortcut,
  title={Shortcut learning susceptibility in vision classifiers},
  author={Suhail, Pirzada and Goel, Vrinda and Sethi, Amit},
  journal={arXiv preprint arXiv:2502.09150},
  year={2025}
}

@inproceedings{long2025seeing,
  title={{Seeing Eye to AI? Applying Deep-Feature-Based Similarity Metrics to Information Visualization}},
  author={Long, Sheng and Chatzimparmpas, Angelos and Alexander, Emma and Kay, Matthew and Hullman, Jessica},
  booktitle={Proceedings of the 2025 CHI Conference on Human Factors in Computing Systems},
  pages={1--20},
  year={2025}
}

@inproceedings{proma2025evaluating,
  title={Evaluating Line Chart Strategies for Mitigating Density of Temporal Data: The Impact on Trend, Prediction, and Decision-Making},
  author={Proma, Rifat Ara and Quadri, Ghulam Jilani and Rosen, Paul},
  booktitle={International Symposium on Visual Computing},
  pages={223--235},
  year={2025},
  organization={Springer}
}

@article{foumani2024survey,
  author    = {Foumani, Navid Mohammadi and Miller, Lynn and Tan, Chang Wei
               and Webb, Geoffrey I. and Forestier, Germain and Dempster, Angus},
  title     = {Deep Learning for Time Series Classification and Extrinsic
               Regression: A Current Survey},
  journal   = {ACM Computing Surveys},
  volume    = {56},
  number    = {9},
  pages     = {1--45},
  year      = {2024},
  doi       = {10.1145/3649448},
}

@article{fawaz2019deep,
  author    = {Ismail Fawaz, Hassan and Forestier, Germain and Weber, Jonathan
               and Idoumghar, Lhassane and Muller, Pierre-Alain},
  title     = {Deep Learning for Time Series Classification: A Review},
  journal   = {Data Mining and Knowledge Discovery},
  volume    = {33},
  number    = {4},
  pages     = {917--963},
  year      = {2019},
  doi       = {10.1007/s10618-019-00619-1},
}

@article{dau2019ucr,
  author    = {Dau, Hoang Anh and Bagnall, Anthony and Kamgar, Kaveh and
               Yeh, Chin-Chia Michael and Zhu, Yan and Gharghabi, Shaghayegh
               and Ratanamahatana, Chotirat Ann and Keogh, Eamonn},
  title     = {The {UCR} Time Series Classification Archive},
  journal   = {IEEE/CAA Journal of Automatica Sinica},
  volume    = {6},
  number    = {6},
  pages     = {1293--1305},
  year      = {2019},
  doi       = {10.1109/JAS.2019.1911747},
}

@article{chung2020introduction,
  title={Introduction to logistic regression},
  author={Chung, Moo K},
  journal={arXiv preprint arXiv:2008.13567},
  year={2020}
}

@article{middlehurst2024bakeoff,
  author    = {Middlehurst, Matthew and Sch\"{a}fer, Patrick and Bagnall, Anthony},
  title     = {Bake Off Redux: A Review and Experimental Evaluation of Recent
               Time Series Classification Algorithms},
  journal   = {Data Mining and Knowledge Discovery},
  volume    = {38},
  number    = {4},
  pages     = {1958--2031},
  year      = {2024},
  doi       = {10.1007/s10618-024-01022-1},
}

@inproceedings{wang2015imaging,
  author    = {Wang, Zhiguang and Oates, Tim},
  title     = {Imaging Time-Series to Improve Classification and Imputation},
  booktitle = {Proceedings of the Twenty-Fourth International Joint Conference
               on Artificial Intelligence (IJCAI)},
  pages     = {3939--3945},
  year      = {2015},
}

@inproceedings{hatami2018classification,
  author    = {Hatami, Nima and Gavet, Yann and Debayle, Johan},
  title     = {Classification of Time-Series Images Using Deep Convolutional
               Neural Networks},
  booktitle = {Proceedings of the 10th International Conference on Machine
               Vision (ICMV)},
  pages     = {106960Y},
  year      = {2018},
  doi       = {10.1117/12.2309486},
}

@article{eckmann1987recurrence,
  author    = {Eckmann, Jean-Pierre and Kamphorst, S. Oliffson and Ruelle, David},
  title     = {Recurrence Plots of Dynamical Systems},
  journal   = {Europhysics Letters},
  volume    = {4},
  number    = {9},
  pages     = {973--977},
  year      = {1987},
  doi       = {10.1209/0295-5075/4/9/004},
}

@inproceedings{mariani2024fusion,
  author    = {Mariani, Maria C. and Bhuiyan, Md Amin Al and
               Tweneboah, Osei K. and Gonzalez-Huizar, Hector},
  title     = {Transforming Time Series into Texture Images: A Fusion of
               Recurrence Plots and Gramian Angular Fields},
  booktitle = {Proceedings of the Hawaii University International Conferences
               (HUIC-STEM)},
  year      = {2024},
}

@article{zhao2024cwt3dcnn,
  author    = {Zhao, Bo and Lu, Haijiang and Chen, Shengyong and Liu, Jing
               and Wu, Dazhong},
  title     = {Fault Detection and Identification Method: {3D-CNN} Combined
               with Continuous Wavelet Transform},
  journal   = {Computers \& Chemical Engineering},
  pages     = {108761},
  year      = {2024},
  doi       = {10.1016/j.compchemeng.2024.108761},
}

@article{oh2025tssi,
  author    = {Oh, Young Keun and Kim, Hyunsoo and Kim, Sungzoon},
  title     = {{TSSI}: Time Series as Screenshot Images for Multivariate
               Time Series Classification Using Convolutional Neural Networks},
  journal   = {Computers \& Industrial Engineering},
  year      = {2025},
  doi       = {10.1016/j.cie.2024.110820},
}

@inproceedings{li2023vitst,
  author    = {Li, Zekun and Liu, Fengqi and Yang, Wenjie and Peng, Shiyang
               and Zhou, Jun},
  title     = {Time Series as Images: Vision Transformer for Irregularly
               Sampled Time Series},
  booktitle = {Advances in Neural Information Processing Systems (NeurIPS)},
  year      = {2023},
}

@inproceedings{alghanemi2024cwt,
  author    = {Alghanemi, Salem and others},
  title     = {Enhancing Multivariate Time Series Forecasting Through
               Integration of {CWT} Scalograms as {CNN} Channels},
  booktitle = {Studies in Systems, Decision and Control},
  publisher = {Springer},
  year      = {2024},
  doi       = {10.1007/978-3-031-71649-2_3},
}

@inproceedings{kornblith2019similarity,
  author    = {Kornblith, Simon and Norouzi, Mohammad and Lee, Honglak and Hinton, Geoffrey},
  title     = {Similarity of Neural Network Representations Revisited},
  booktitle = {Proceedings of the International Conference on Machine Learning (ICML)},
  pages     = {3519--3529},
  year      = {2019},
}

@inproceedings{geirhos2019texture,
  author    = {Geirhos, Robert and Rubisch, Patricia and Michaelis, Claudio
               and Bethge, Matthias and Wichmann, Felix A. and Brendel, Wieland},
  title     = {{ImageNet}-trained {CNNs} are Biased Towards Texture; Increasing
               Shape Bias Improves Accuracy and Robustness},
  booktitle = {Proceedings of the International Conference on Learning
               Representations (ICLR)},
  year      = {2019},
}

@article{geirhos2020shortcut,
  author    = {Geirhos, Robert and Jacobsen, J\"{o}rn-Henrik and Michaelis, Claudio
               and Zemel, Richard and Brendel, Wieland and Bethge, Matthias
               and Wichmann, Felix A.},
  title     = {Shortcut Learning in Deep Neural Networks},
  journal   = {Nature Machine Intelligence},
  volume    = {2},
  pages     = {665--673},
  year      = {2020},
  doi       = {10.1038/s42256-020-00257-z},
}

@inproceedings{selvaraju2017gradcam,
  author    = {Selvaraju, Ramprasaath R. and Cogswell, Michael and Das, Abhishek
               and Vedantam, Ramakrishna and Parikh, Devi and Batra, Dhruv},
  title     = {{Grad-CAM}: Visual Explanations from Deep Networks via
               Gradient-Based Localization},
  booktitle = {Proceedings of the IEEE International Conference on Computer
               Vision (ICCV)},
  pages     = {618--626},
  year      = {2017},
  doi       = {10.1109/ICCV.2017.74},
}

@inproceedings{selvaraju2019hint,
  author    = {Selvaraju, Ramprasaath R. and Lee, Stefan and Shen, Yilin
               and Jin, Hongxia and Ghosh, Shalini and Heck, Larry
               and Batra, Dhruv and Parikh, Devi},
  title     = {Taking a {HINT}: Leveraging Explanations to Make Vision and
               Language Models More Grounded},
  booktitle = {Proceedings of the IEEE/CVF International Conference on
               Computer Vision (ICCV)},
  pages     = {2591--2600},
  year      = {2019},
}

@article{mcinnes2018umap,
  author    = {McInnes, Leland and Healy, John and Saul, Nathaniel and
               Gro{\ss}berger, Lukas},
  title     = {{UMAP}: Uniform Manifold Approximation and Projection for
               Dimension Reduction},
  journal   = {Journal of Open Source Software},
  volume    = {3},
  number    = {29},
  pages     = {861},
  year      = {2018},
  doi       = {10.21105/joss.00861},
}

@book{jolliffe2002principal,
  author    = {Jolliffe, Ian T.},
  title     = {Principal Component Analysis},
  year      = {2002},
  publisher = {Springer},
  address   = {New York},
  edition   = {2nd}
}

@inproceedings{kingma2015adam,
  author    = {Kingma, Diederik P. and Ba, Jimmy},
  title     = {Adam: A Method for Stochastic Optimization},
  booktitle = {Proceedings of the International Conference on Learning
               Representations (ICLR)},
  year      = {2015},
}

@article{venkatesan2026vtbenchmultimodalframeworktimeseries,
  author       = {Madhumitha Venkatesan and
                  Xuyang Chen and
                  Dongyu Liu},
  title        = {VTBench: {A} Multimodal Framework for Time-Series Classification with
                  Chart-Based Representations},
  journal      = {CoRR},
  volume       = {abs/2604.27259},
  year         = {2026}
}

@article{alain2017probing,
  title={Understanding intermediate layers using linear classifier probes},
  author={Alain, Guillaume and Bengio, Yoshua},
  journal={ICLR Workshop},
  year={2017}
}

@article{hunter2007matplotlib,
  author    = {Hunter, John D.},
  title     = {Matplotlib: A 2D Graphics Environment},
  journal   = {Computing in Science \& Engineering},
  volume    = {9},
  number    = {3},
  pages     = {90--95},
  year      = {2007},
  doi       = {10.1109/MCSE.2007.55},
}

@article{mallat1989wavelet,
  author    = {Mallat, Stephane G.},
  title     = {A Theory for Multiresolution Signal Decomposition:
               The Wavelet Representation},
  journal   = {IEEE Transactions on Pattern Analysis and Machine Intelligence},
  volume    = {11},
  number    = {7},
  pages     = {674--693},
  year      = {1989},
  doi       = {10.1109/34.192463},
}

@article{griffin1984stft,
  author    = {Griffin, Daniel W. and Lim, Jae S.},
  title     = {Signal Estimation from Modified Short-Time {Fourier} Transform},
  journal   = {IEEE Transactions on Acoustics, Speech, and Signal Processing},
  volume    = {32},
  number    = {2},
  pages     = {236--243},
  year      = {1984},
  doi       = {10.1109/TASSP.1984.1164317},
}

@inproceedings{islam2021shape,
  author       = {Md. Amirul Islam and
                  Matthew Kowal and
                  Patrick Esser and
                  Sen Jia and
                  Bj{\"{o}}rn Ommer and
                  Konstantinos G. Derpanis and
                  Neil D. B. Bruce},
  title        = {Shape or Texture: Understanding Discriminative Features in CNNs},
  booktitle    = {9th International Conference on Learning Representations, {ICLR} 2021,
                  Virtual Event, Austria, May 3-7, 2021},
  publisher    = {OpenReview.net},
  year         = {2021}
}

\end{document}